\documentclass{svproc}
\usepackage{url}

\usepackage{hyperref}       % hyperlinks
\usepackage{graphicx}
\usepackage{float}
\usepackage{makecell}
\usepackage{lscape}
\usepackage{booktabs}       % professional-quality tables
\usepackage[utf8]{inputenc} % allow utf-8 input
\usepackage[T1]{fontenc}    % use 8-bit T1 fonts
\usepackage[nolist]{acronym}

\usepackage{varwidth}
\newsavebox\tmpbox
\usepackage{subcaption}
\usepackage[style=numeric,sorting=none,maxbibnames=9]{biblatex}
\DeclareFieldFormat{labelnumberwidth}{#1.}

\begin{document}
\mainmatter              % start of a contribution
%
%Inser your acronyms here
\begin{acronym}[UML]
  \acro{LGS}{Light Guided Systems}
  \acro{AI}{Artificial Intelligence}
  \acro{API}{Application Programming Interface}
  \acro{RGB}{Red, Green, Blue Color Model}
  \acro{GDPR}{General Data Protection Regulation}
  \acro{RGB-D}{Red, Green, Blue Color Model and Corresponding Depth Model}
  \acro{RGB-D IR}{Red, Green, Blue Color Model and Corresponding Infrared Depth Model}
  \acro{HAR}{Human Action Recognition}
  \acro{HAP}{Human Action Prediction}
  \acro{EYTH}{EgoYouTubeHands}
  \acro{RPN}{Region Proposal Network}
  \acro{MPL}{Multi Product Line}
  \acro{NLP}{Natural Language Processing}
  \acro{GTN}{Gated Transformer Networks}
  \acro{POC}{Proof of Concept}
  \acro{IMU}{Inertial Measurement Unit}
  \acro{LRCN}{Long-term Recurrent Convolutional Network}
  \acro{CI}{Continuous Integration}
  \acro{CD}{Continuous Deployment}
  \acro{IIoT}{Industrial Internet of Things}
  \acro{GPU}{Graphical Processing Unit}
  \acro{ConvNet}{Convolutional Neural Network}
  \acro{MiCT}{Mixed 3D/2D Convolutional Tube}
  \acro{RNN}{Recurrent Neural Network}
  \acro{LSTM}{Long Short Term Memory}
  \acro{HHI}{Human-Human Interaction}
  \acro{HOI}{Human-Object Interaction}
  \acro{SAS}{Shared Access Signature}
  \acro{SSD}{Single Shot Detector}
  \acro{ST-CNN}{Spatial-Temporal Convolutional Neural Network}
  \acro{TP}{True Positives}
  \acro{FP}{False Positives}
  \acro{TN}{True Negatives}
  \acro{FN}{False Negatives}
\end{acronym}
\title{Challenges of the Creation of a Dataset for Vision Based Human Hand Action Recognition in Industrial Assembly}
\titlerunning{Dataset for Human Hand Action Recognition}  % abbreviated title (for running head)
\author{Fabian Sturm\inst{1} \and Elke Hergenroether\inst{2} \and Julian Reinhardt\inst{1}
\and Petar Smilevski Vojnovikj\inst{1} \and Melanie Siegel\inst{2}}
\authorrunning{Fabian Sturm et al.} % abbreviated author list (for running head)
\tocauthor{Fabian Sturm, Elke Hergenroether, Julian Reinhardt, Petar Smilevski Vojnovikj, Melanie Siegel}
\institute{Bosch Rexroth AG, Lise-Meitner-Strasse 4, 89081 Ulm, Germany
\\
\email{fabian.sturm@bosch.com}%,\\ WWW home page:
%\texttt{http://users/\homedir iekeland/web/welcome.html}
\and
University of Applied Sciences Darmstadt, Schoefferstraße 3,
64295 Darmstadt, Germany}
\maketitle              % typeset the title of the contribution
\begin{abstract}
This work presents the Industrial Hand Action Dataset V1, an industrial assembly dataset consisting of 12 classes with 459,180 images in the basic version and 2,295,900 images after spatial augmentation. Compared to other freely available datasets tested, it has an above-average duration and, in addition, meets the technical and legal requirements for industrial assembly lines. Furthermore, the dataset contains occlusions, hand-object interaction, and various fine-grained human hand actions for industrial assembly tasks that were not found in combination in examined datasets. The recorded ground truth assembly classes were selected after extensive observation of real-world use cases. A Gated Transformer Network, a state-of-the-art
model from the transformer domain was adapted, and proved with a test accuracy of 86.25\% before hyperparameter tuning by 18,269,959 trainable parameters, that it is possible to train sequential deep learning models with this dataset.
\end{abstract}
\section{Introduction}
\label{Introduction}
The full automation of production processes in industrial assembly lines has shown that humans cannot be completely replaced by machines. This is mainly for monetary reasons, such as high fixed maintenance costs for work tasks that are too complex for robotics and other machines and can currently only be solved economically by humans. The disadvantage for humans, resulting from the increasing variance of products and the desired flexibility of the worker in assembly, is the susceptibility of humans to errors, which increases due to the increased workload. This is caused primarily by decreasing concentration during long shifts, inattention, or distraction. Increasing demands also mean that assembly workers need more and more knowledge to assemble products correctly and therefore have less time for training and familiarization, resulting in a lack of experience. These initially undetected assembly errors then continue through the production process until they are only noticed in quality control or even when the product is in operation at the customer's site, with unpleasant consequences for production, such as an increased reject rate or damage to the company's image due to the low quality of the products.\\
To counteract these weaknesses of manual assembly by humans, ensure high product quality and minimize the error rate, an intelligent assistance system that recognizes the actions of the assembler by checking the assembly steps and provides direct feedback so that the assembler can immediately correct the mistake made is urgently needed. The smart assistance system can be provided by deep learning approaches via the visual recognition of human actions. However, in deep learning, examples are needed to train the network architectures appropriate for the task to find patterns and correlations in the data for a final classification. Since industrial assembly is performed by hand, the datasets studied in this work are mostly based on the movement of human hands and the interaction between tools and objects in the red, green, blue color model (RGB) and red, green, blue color model
and corresponding depth model (RGB-D). Based on the weaknesses of the presented datasets for industrial usability, a specially created dataset is presented that meets the industrial requirements and proves that the created dataset can be used for training a deep learning model. The requirements placed on the dataset in this context are to recreate an industrial scenario in great detail, taking into account legal data protection regulations. Therefore, the focus is exclusively on hands. Furthermore, the dataset should contain occlusions of the hands by e.g., bigger parts, clear finely granular actions as well as object interactions and no non-industrial disturbances by the environment, like reaching for a smartphone or similar objects which are not part of the assembly process in order to be considered as ground truth for the training of a deep learning model. Additionally, since the focus relies on human hands and to avoid feature extraction problems from the environment, the video sequences are for the model training preprocessed into extracted features by a hand pose estimation approach to focus the network only on the behavior of a skeleton of human hands. 
The remaining work is structured as follows. In Sect. \ref{Relateddatasets} open sourced datasets which are also part of a scientific publication for \ac{HAR} and gesture control are examined. Sect. \ref{Weaknessesofexistingdatasets} deals with the analysis of these datasets and explains the usability aspects for the desired industrial application. In Sect. \ref{AssemblyRecordingSetup} the recording setup is described before the introduction of the created 'Industrial Hand Action Dataset V1' in Sect. \ref{HandAssemblyDataset}. Sect.\ref{Modeltraining}, explains the used network architecture used to test trainability on the dataset, followed by the description of the training environment. In Sect. \ref{Results} the results of the creation of the dataset are examined, followed by planned future tasks and a final conclusion in Sect. \ref{Conclusion}.
\section{Related Datasets}
\label{Relateddatasets}
\vspace{-10mm}
\begin{table}[H]
  \caption{Human Action Recognition Dataset Comparison}
  \label{table:shortdatasetcomparisontable}
  \centering
  \resizebox{\textwidth}{!}{\begin{tabular}{cccccc}
    \toprule
    \cmidrule(r){1-6}
    Dataset & Classes & Activity & Environment & \makecell{GDPR\\Constrains} & View\\
    \midrule
    IPN Hand \cite{DBLP:journals/corr/abs-2005-02134} & 14 & \makecell{Human Computer\\Interaction} & Open Space & Yes & Third Person\\
    EgoHands+ \cite{EgoHands} & 16 & Playing Games & \makecell{Office/Courtyard/\\Family Room} & Yes & First Person\\
    EYTH (EgoYouTubeHands) \cite{aisha} & X & \makecell{Hand Object\\ Interaction} & Daily Live & Yes & \makecell{First/\\Third Person}\\
    Ikea Assembly Dataset \cite{IKEAASM} & 4	& Assembly	& Office/Family Room & Yes & Third Person\\
    20bn-something-something-v2 \cite{Goyal2017} & 174	& \makecell{Hand Object\\ Interaction} & Daily Live & No & First Person\\
    Cambridge Hand\\ Gesture Dataset \cite{tensorcanonical} & 9	& Gestures	& Clean Background	& No & First Person\\
    EGTEA Gaze+ \cite{egtea} & 106	& Cooking & Kitchen	& No & First Person\\
    MPII Cooking 2 \cite{Rohrbach2016} & 59	& Cooking	& Kitchen	& Yes & First Person\\
    YouCook 2 \cite{Zhou2018} & 89	& Cooking & Kitchen	& Yes & Third Person\\
    COIN \cite{Tang2019} & 180	& \makecell{Hand Object\\ Interaction}	& Daily Live & Yes & \makecell{First/\\Third Person}\\
    Drive\&Act \cite{driveandact} & 83	& Driving	& Driver Cabine	& Yes & Third Person\\
    WorkingHands \cite{workinghands} & 13	& Assembly	& Workbench	& No & First Person\\
    ChaLearn Iso/ConGD \cite{chalearn} & 249 & Gestures	& Open Space & Yes & Third Person\\
    LTTM Senz3D \cite{lttm1}\cite{lttm2} & 11 & Gestures	& Open Space & Yes & Third Person\\
    Multi-Modal Hand\\ Activity Video Dataset \cite{multi-modal_hands} & 15	& \makecell{Hand Object\\ Interaction}	& Workbench	& No & First Person\\
    EgoGesture Dataset \cite{Zhang2018} & 83	& Gestures	& Indoor/Outdoor	& No & First Person\\
    Gun71 \cite{rogez} & 71	& Grasping	& Family Room	& No & First Person\\
    \bottomrule
  \end{tabular}}
%\raggedright{X = Information not specified in reference}
\end{table}
\vspace{-5mm}
Human actions often appear in data recorded by visual sensors, more specifically \ac{RGB} or \ac{RGB-D} cameras, in the form of image or video data \cite{DBLP:journals/corr/abs-2201-05761}. One of the reasons for this visual type of data acquisition is that they are rich in features that can be used to remove ambiguities after preprocessing. As can be seen from the existing literature, extensive research has been done in the last decade, especially in the field of video classification as the important works for \ac{HAR} by \cite{Ji2013}\cite{Feichtenhofer2016}\cite{Carreira2017} proves. A classical approach in this regard for \ac{HAR} is to extract features from individual 2D frames of \ac{RGB} videos and use these features to train models to classify human actions.
For this purpose, datasets already exist that are used to provide models with the necessary examples to recognize specific human actions. These examined datasets contain human assembly tasks, \cite{IKEAASM}\cite{workinghands} hand gestures for human computer interaction, \cite{DBLP:journals/corr/abs-2005-02134} 
cooking tasks, \cite{Zhou2018}\cite{Singh2016} which are similar to \ac{HOI} in the preparation of the ingredients or \ac{HOI} in tasks like playing games \cite{EgoHands}.
Furthermore, it is important to mention that these existing and publicly available datasets were created under specific environmental conditions, such as in free environment \cite{aisha}\cite{Tang2019} in closed rooms \cite{rogez}\cite{IKEAASM} or under laboratory conditions \cite{tensorcanonical}\cite{multi-modal_hands}. In the following, datasets are presented that are suited for especially industrial and real-world use cases. They are divided into the domains of assembly tasks \ref{Assemblydatasets}, gestures \ref{Gesturesdatasets}, hand-object interactions \ref{handobjectinteractionsdataset}, cooking \ref{Cookingdatasets}, and daily tasks \ref{Dailytasks}. All the following information can be viewed in a summarized form in Table \ref{table:shortdatasetcomparisontable} or in a detailed version regarding the technical specification in the Appendix \ref{Appendix} in Table \ref{table:datasetcomparisontable}. While it should be noted that some divisions of the data sets are not entirely clear, an attempt has been made to divide the datasets as much as possible. The goal in the next chapter is to gain a better understanding of fine-grained human actions, with focus on industrial environments.\\
\subsection{Assembly Datasets}
\label{Assemblydatasets}
%
%Assembly
A wide variety of different assembly types and different time scales is provided by the Ikea Assembly Dataset \cite{IKEAASM}. In this approach, the assembly of furniture is done with the same type of components in several ways. It consists of natural, as well as unusual, human poses which are visually very similar and contain fine-grained \ac{HOI}'s. Data were collected from three different camera views to process the complete human body, object, and self occlusions from multiple sensor modalities, including color, depth, and surface normal.
%WorkingHands
Comparable to the Ikea Assembly Dataset is the Working Hands dataset, \cite{workinghands} which was built for manufacturing tasks. The difference to the above-mentioned one is that it was created to focus only on the hands and interaction between tools and parts. To get a wide range of examples, the Working Hands dataset is a mixture of real and synthetic interaction data. Therefore, not only the tools but also hands were created synthetically with the goal to segment the hands and tools for interaction.
%Multi-Modal Hand Activity Video Dataset
Similar manufacturing \ac{HOI} are provided in the Multi-Modal Activity Video Dataset, \cite{multi-modal_hands} in which pixel-wise hand segmentation is used to segment hands from objects by thermal and \ac{RGB-D} cameras. 
Besides, datasets that were created only for assembly or manufacturing tasks, human gesture recognition can also take into account to recognize human actions, especially if the hands are the only recorded body part.
\subsection{Gestures Datasets}
\label{Gesturesdatasets}
%
%ChaLearn Iso/ConGD
A very early approach was made by \cite{chalearn} with the ChaLearn Iso/ConGD dataset, \cite{chalearn} consisting of 54,000 hand and arm gestures recorded with a \ac{RGB-D} camera for 249 classes in 22,535 manually labeled videos from a third person view. The videos are organized into 100 gestures belonging to a small gesture vocabulary of 8 to 12 gestures recorded by the same user. In addition, there is a subset of augmented batches in which the horizontal position of the user is randomly shifted or scaled. A similar recorded dataset is the LTTM Senz3D \cite{lttm1}\cite{lttm2}. This dataset was recorded with gestures performed by 4 different people, performing 11 different gestures and repeated 30 times, for a total of 1,320 samples.
A later created approach and one of the leading datasets for the human gesture
recognition is the IPN Hands dataset, \cite{DBLP:journals/corr/abs-2005-02134} that was made for human computer interactions via touchless screens. This dataset was recorded in a third-person view in open spaces. It contains 4,000 clips of \ac{RGB} data, a high amount of videos comparatively to the other inspected datasets in this work. The dataset offers static but also dynamic gestures with considerable variation, as well as clutter backgrounds, strong and weak illumination conditions and static and dynamic background environments. %EgoGesture & Tensor Canonical Correlation Analysis for Action Classification
However, the focus of the gesture recognition datasets are not only on clean hand gestures, as is the case with the Cambridge Hand Gesture Dataset \cite{tensorcanonical}. These datasets may also include gestures from scenes with \ac{HOI}.
\subsection{Human Object Interaction Datasets}
\label{handobjectinteractionsdataset}
The EgoGesture dataset \cite{Zhang2018} consists of 83 different static and dynamic egocentric gestures, focused on interaction with wearable devices. Over 24,000 video sequences were recorded in six different indoor and outdoor scenes with different backgrounds and lighting. Therefore, the data is highly diverse. The goal was to segment gestures in continuous data, and thus to be able to evaluate different approaches for gesture detection and classification in numerous ways.
%COIN
An instructional video dataset for daily life tasks comparable to assembly instructions is the COIN dataset \cite{Tang2019}. The creators had a similar goal to the creators of the EgoGesture dataset, which was to create a collection of various real-world activities but from YouTube videos. The dataset consists of 11,827 videos related to 180 instructional tasks from several domains, and each step is connected to a label. This structure is also the difference to other instructional video datasets because it is organized in a three-level semantic structure \cite{Tang2019}.\\
%%%%
%playing games
%EgoHands
Gestures and assembly tasks are very obvious to describe human behavior in certain scenes, but also the interaction between humans, hands and hand-objects has to be considered.
The EgoHands dataset, \cite{EgoHands} contains individuals out of an egocentric view in 48 different videos of dynamic \ac{HOI} within playing games with pixel-level ground-truth annotations for 4,800 frames and ground truth segmentation masks for over 15,000 hands.
Beside the \ac{HOI}, the dataset includes more realistic and challenging social situations where, different from the other datasets, multiple sets of hands appear in the view. %20bn something something
In contrast to previous datasets, the description of a single action in the 20bn-something-something v2 dataset, \cite{Goyal2017} was made by a natural language template to present the labels instead of a fixed data structure to make the human action's description more fine-granular \cite{Goyal2017}. With 108,499 video clips and 174 classes, this dataset is one of the biggest dataset currently existing. The recording of crowd workers was mostly based on \ac{HOI}.
%EGTEA Gaze+
But not only tasks like playing games or similar can be used as examples for human actions. Also, human hand movements during cooking, which are very similar to movements in industrial assembly lines, were recorded. 
\subsection{Cooking Datasets}
\label{Cookingdatasets}
A good example is the large EGTEA Gaze+ dataset, which is an extended version of the former GTEA dataset \cite{egtea}. It was created for egocentric views with wearable cameras on cooking tasks. As a result, the very specific, fine-grained human actions were used for segmentation and classification.
%MPII Cooking 2 & %Youcook2
Similar to the forementioned dataset with fine-grained cooking tasks is the MPII Cooking 2 \cite{Rohrbach2016}, which was live recorded from a front view and the Youcook 2 which is a collection of cooking videos from several views from YouTube \cite{Zhou2018}, with the goal to differentiate between fine-grained body motions and the categories of the tasks.
\subsection{Daily Tasks Datasets}
\label{Dailytasks}
%
%EgoYouTubeHands (EYTH)
A collection of several daily task videos from YouTube, recorded from an egocentric view, is the \ac{EYTH} dataset by \cite{aisha}. The goal was to create a "hands-in-the-wild" dataset for detection and segmentation. Regarding the authors, the limitation of existing hand action datasets are caused by laboratory settings and can therefore not be used in open worlds. The goal for this dataset is to detect any hand, especially in first person videos recorded in unconstrained daily settings.
\\
%Gun71
Specialized on grasps is the Grasp Understanding dataset also known as Gun71 with 12,000 \ac{RGB-D} images of scenes from hand to object manipulations in typical house scenes from a chest-mounted \ac{RGB-D} camera from 5 to 6 views each \cite{rogez}.
%Drive%Act
A different kind of use case for \ac{HAR} was pursued in the Drive\&Act dataset \cite{driveandact}. The goal was to use the dataset to recognize fine-grained human behavior inside the vehicle cabin. Compared to the other datasets, the Drive\&Act was taken from 6 different angles with \ac{RGB-D IR} information, aiming to see the location, the objects for passenger interaction and two types of labels for the actual action. With more than 9.6 million images, this is the largest dataset presented in this paper. 

%Deeper technical insights are documented in Table \ref{table:datasetcomparisontable} where the evaluation criteria of the datasets were selected from the literature based on existing evaluations.
%
\section{Weaknesses of Existing Datasets for Industrial Assembly Lines and Real World Applications}
\label{Weaknessesofexistingdatasets}
The computer vision approach brings some problems which are only partly covered in the existing \ac{HAR} datasets from Sect. \ref{Relateddatasets}. This concerns technical aspects as well as aspects specific to industrial environments. %These aspects are listed in Table \ref{table:shortdatasetcomparisontable} as well as in a detailed version in the Appendix \ref{Appendix} in Table \ref{table:datasetcomparisontable}.
\subsection{Technical Weaknesses}
\label{technicalrequirements}
\textbf{Occlusions}\\
Technical aspects that mainly affect the recognition and prior training of this approach are the occlusion of the hands and work steps by objects, more precisely by components which obscure significant recognition features of the hand. Since this can only happen due to \ac{HOI} it is not covered by IPN Hand \cite{DBLP:journals/corr/abs-2005-02134}, Cambridge Hand Gesture Dataset \cite{tensorcanonical}, ChaLearn Iso/ConGD \cite{chalearn}, LTTM Senz3D \cite{lttm1}\cite{lttm2} and the EgoGesture Dataset \cite{Zhang2018}.\\
\\\textbf{Depth Data}\\
The small distance between the work surface and the worker's hands, which makes hand detection difficult, is also an occurrence that must be taken into account and is not fully captured in the data sets studied. Many \ac{HAR} approaches use depth data that should provide the necessary spatial information of human behavior as input to a deep learning network, like the Ikea Assembly Dataset \cite{IKEAASM}, the WorkingHands Dataset \cite{workinghands}, ChaLearn Iso/ConGD \cite{chalearn}, LTTM Senz3D \cite{lttm1}\cite{lttm2} and GUN71 \cite{rogez}. However, the data were mostly recorded in open spaces, such as in front of a green screen in the Cambridge Hand Gesture Dataset \cite{tensorcanonical}, rather than in industrial or unsupervised conditions with a lot of noise in the foreground and background, \cite{DBLP:journals/corr/abs-2005-02134}\cite{chalearn}\cite{lttm1}\cite{lttm2}, which is a weakness in scaling the trained network. These occurrences in the environment due to tools or parts that need to be assembled can negatively affect the depth data. Also, the close distance between the depth camera and the worker to the work surface can result in an image with a lot of noise in the data \cite{Jauch2021}.\\
\\\textbf{Bias}\\
Furthermore, the classes and tasks of the test person are sometimes unclear like in the EYTH dataset, \cite{aisha}, or have lots of environmental bias like the YouCook2 dataset \cite{Zhou2018}. This bias can also be caused by more than just one person in the camera view, like in the EgoHands+ dataset, where several people appear \cite{EgoHands}.
\vspace{-3mm}
\subsection{Legal Requirements}
\label{legalrequirements}
In addition to the technical requirements that need to be considered in order to create a dataset that is suitable for industrial use, the non-technical aspects more precise the legal requirements also need to be taken into account. In countries of the European Union that have agreed to the \ac{GDPR} or in Germany to the “BDSG-neu” \cite{BDSG} it is necessary to observe the data protection regulations and personal rights of the employees. In countries with a worker's committee, the council also strictly controls what happens to the data when employees are filmed on company premises. The reason for this is that in most cases the recording and tracking of personal data is not permitted under the above-mentioned data protection laws, or only under very strict conditions. This includes the prohibition of recording personal data such as the batch with batch number, the name of the employee on a name tag, the face or other information that leads to the unique identification of the employee. Furthermore, in countries like Germany, it is not permitted that the work task performed can be traced back to the respective employee \cite{BDSG}. Constrains regarding these private data regulation laws are recognized by several datasets where the face or the complete body of the probands are visible like in the IPN Hands dataset \cite{DBLP:journals/corr/abs-2005-02134}  the EgoHands+ \cite{EgoHands}, EYTH \cite{aisha} but also the behavior in the environment without doing the actual task is recorded like in the Ikea Assembly dataset \cite{IKEAASM} or the Drive\&Act dataset \cite{driveandact}.
\subsection{Conclusion of the Weak Points}
\label{conclusionoftheweakpoints}
Since none of these general working conditions and privacy laws are fully met by the freely available existing datasets from Sect. \ref{technicalrequirements} and Sect. \ref{legalrequirements} such as IPN Hand \cite{DBLP:journals/corr/abs-2005-02134}, EgoHands+ \cite{EgoHands}, EYTH \cite{aisha}, Ikea Assembly Dataset \cite{IKEAASM}, 20bn-something-something-v2 \cite{Goyal2017}, Cambridge Hand Gesture Dataset \cite{tensorcanonical}, EGTEA Gaze+ \cite{egtea}, MPII Cooking 2 \cite{Rohrbach2016}, YouCook 2 \cite{Zhou2018}, COIN \cite{Tang2019}, Drive\&Act \cite{driveandact}, WorkingHands \cite{workinghands}, ChaLearn Iso/ConGD \cite{chalearn}, LTTM Senz3D \cite{lttm1}\cite{lttm2}, Multi-Modal Hand Activity Video Dataset \cite{multi-modal_hands},
EgoGesture Dataset \cite{Zhang2018}, Gun71 \cite{rogez}, it is urgent necessary to create a specific dataset for a human hand action recognition approach in industrial assembly lines. This dataset must meet all technical aspects mentioned such as finely granular actions, \ac{HOI}, occlusion by parts, no depth data, no distortions due to environmental disturbances or those that could unnaturally influence the respective action sequence, as well as legal requirements such as no GDPR-critical data. At this point, it has to be noted that many more datasets exists than those mentioned so far, but it was concluded that these listed datasets are the most relevant for comparison with industrial assembly use cases.
%\newpage
%
\vspace{-3mm}
\section{Assembly Recording Setup}
\label{AssemblyRecordingSetup}
\vspace{-5mm}
\begin{figure}[ht]
\centering
\includegraphics[clip,trim=2cm 5cm 2cm 5.5cm,width=0.7\textwidth]{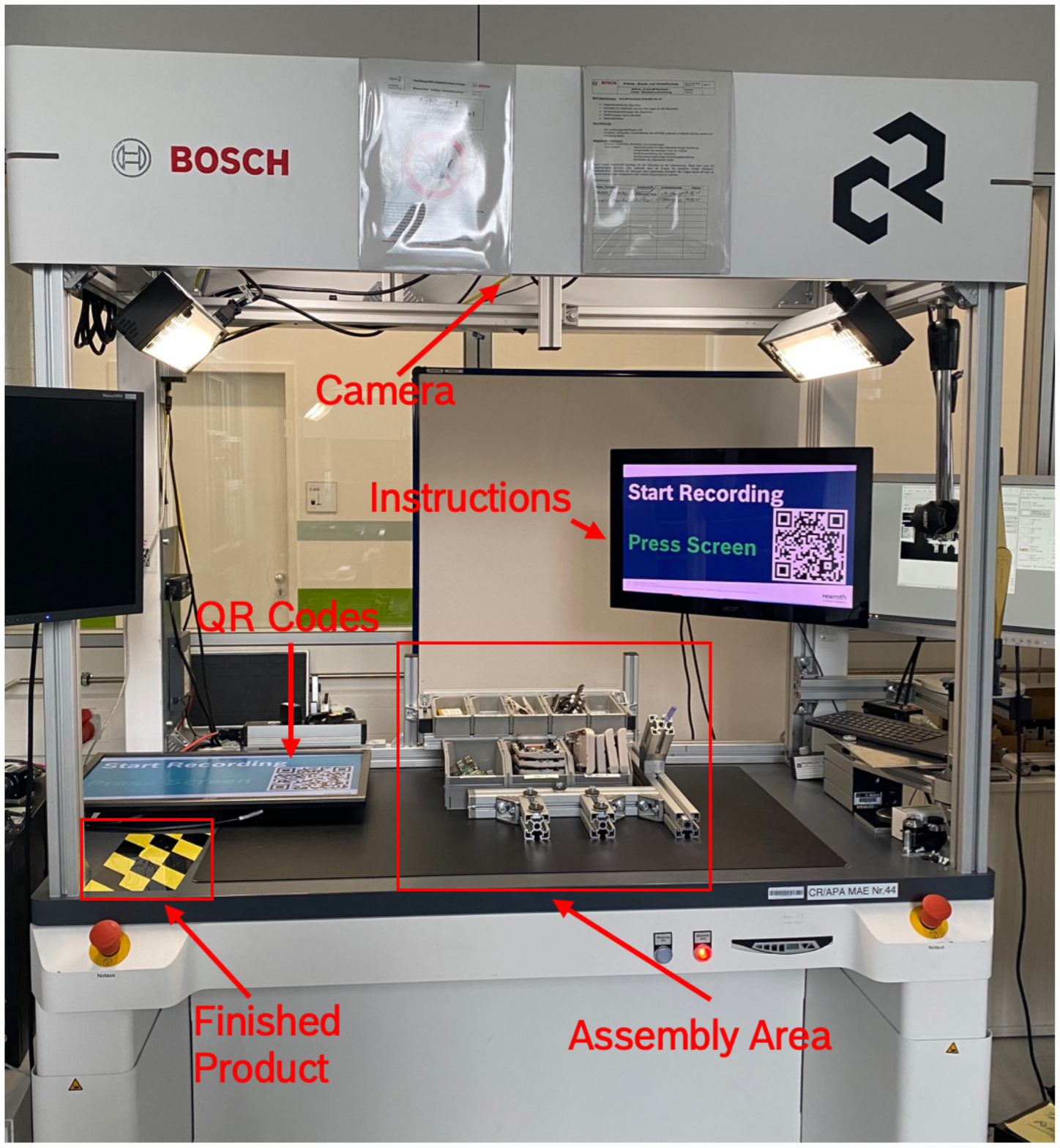}
\caption{Recording Test Bench Front}
\label{testbenchfront}
\end{figure}
\begin{figure}[ht]
\centering
%\vspace{0.35\textwidth}
\includegraphics[clip,trim=5cm 0cm 5cm 0cm,width=0.7\textwidth]{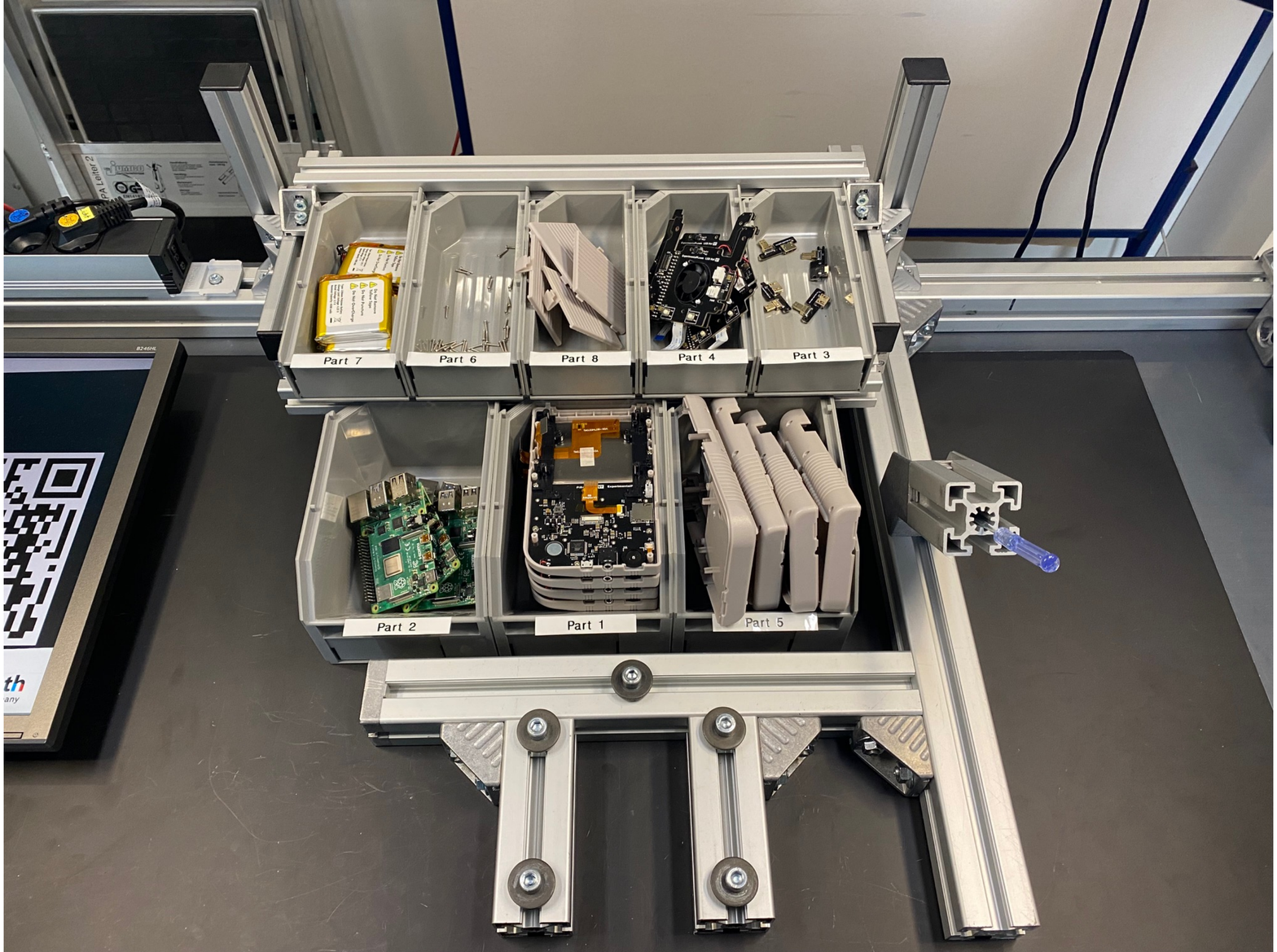}
\caption{Recording Test Bench Mount}
\label{testbenchmount}
\end{figure}

\vspace{-3mm}
In order to have an industrial ground truth dataset, the first step is to create a controllable environment specifically for labeling assembly sequences, but it is also necessary to be as close as possible to the real world use case to prove scalability in a real scenario. Therefore, a PiBoy DMG\footnote{\href{https://experimentalpi.com/PiBoy\-DMG\-\-Kit\_p\_18.html}{https://experimentalpi.com/PiBoy\-DMG\-\-Kit\_p\_18.html}}, see Fig. \ref{fig:PiBoy}, is assembled as an example product for this use case, which is very similar to the real-world scenario under consideration in terms of the assembly procedure.
\begin{minipage}{\textwidth}
  \begin{minipage}{0.49\textwidth}
    \centering
    \includegraphics[clip,trim=6cm 7cm 6cm 6cm,width=0.4\textwidth]{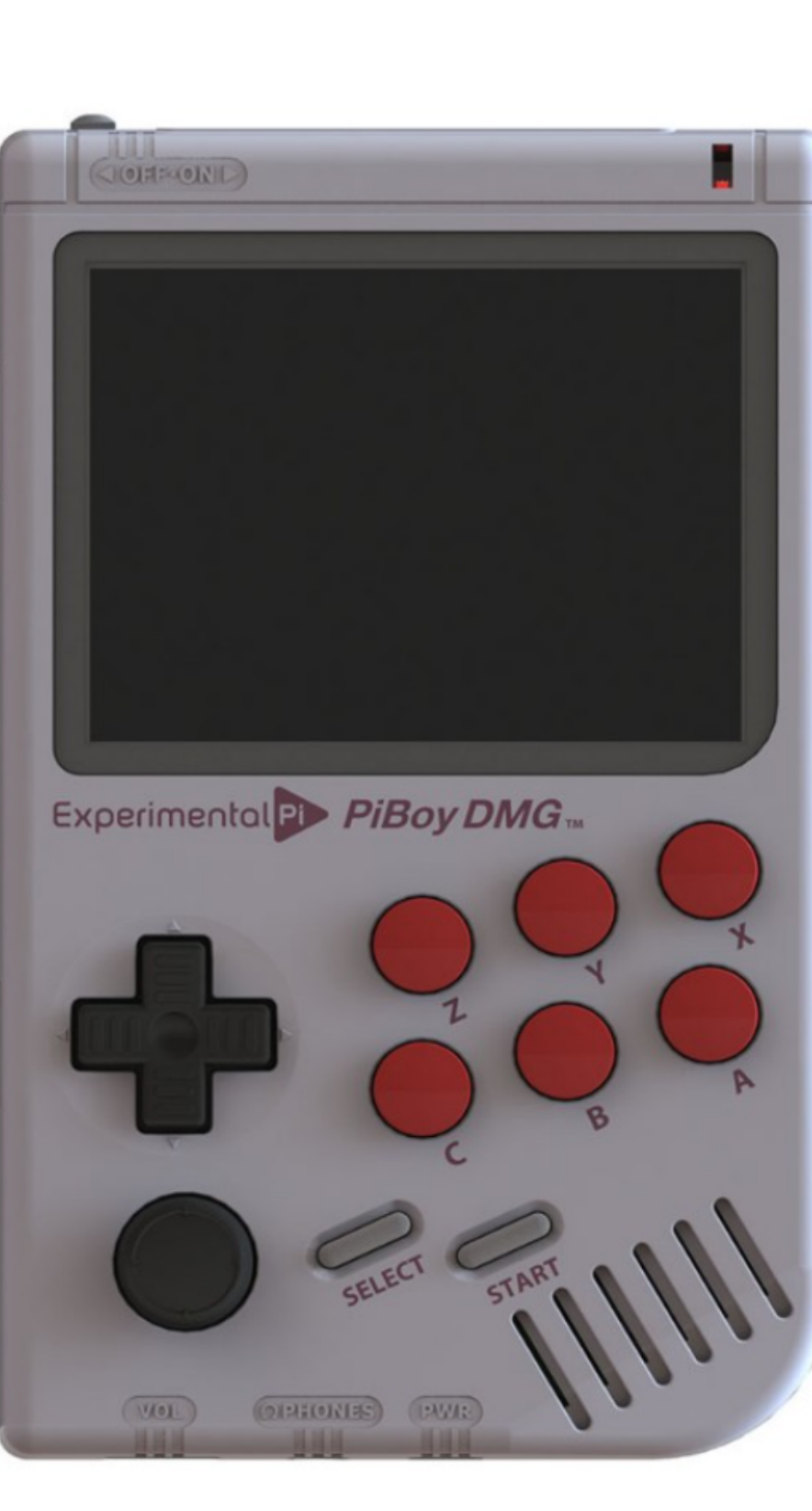}
    \captionof{figure}{Example Product PiBoyDMG}
    \label{fig:PiBoy}
  \end{minipage}
  \hfill
  \begin{minipage}{0.49\textwidth}
    \centering
            \captionof{table}{Part Labels}
            \label{table:parts}
            \begin{tabular}{ll} 
                \toprule
                \cmidrule(r){1-2}
                Part Labels & Product Parts\\
                \midrule
                Part1 & Front Housing\\
                Part2 & Raspberry Pi\\
                Part3 & HDMI Connector\\
                Part4 & Cooling Fan\\
                Part5 & Back Housing\\
                Part6 & Screws\\
                Part7 & Battery\\
                Part8 & Battery Cover\\
                \bottomrule
            \end{tabular}
    \end{minipage}
  \end{minipage}
The product consists in this use case of eight different parts, see Table \ref{table:parts} which are stored in labeled bins on the worktop and need to be assembled in a special mount under the bins on the table, see Fig. \ref{testbenchmount}.
All parts are assembled by hand, and the full assembly process is complete as soon as the assembled product is removed from the camera field of view. The camera is mounted 0.90 meters above the worktop and has a frontal view onto the worktop. As mentioned in Sect. \ref{Weaknessesofexistingdatasets} for the recording, it is mandatory to prove that the camera records no personal information of the assembler, therefore the view of the camera is restricted to only see the worktop more precise the hands and the wrist.
The recording happened with a UI-326xCP-C Camera\footnote{\href{https://en.ids-imaging.com/store/ui-3260cp-rev-2.html}{https://en.ids-imaging.com/store/ui-3260cp-rev-2.html}} from IDS with a resolution of 1936×1216 in \ac{RGB} format with 24fps and each frame is stored as .png datatype.
The labeling and cutting of the different actions is performed by the assembler with the help of QR codes, created and read with OpenCV, which are presented on a second screen, see Fig. \ref{testbenchfront}, to the camera. The QR codes contain the information of the particular action that is performed. These work tasks are each started by clicking on the instruction screen slightly to the right of the worker's field of view, see Fig. \ref{testbenchfront}, and during the recording each frame is saved in an ascending folder structure associated with the label. The assembler stops saving the relevant sequences to the respective classes by clicking on the instruction screen again when the task is finished. Between each work task, the information about the next work step on the instruction screen, is also labeled with a QR code to omit such sequences for the final dataset.
\section{Industrial Hand Assembly Dataset V1}
\label{HandAssemblyDataset}
\begin{figure}
\centering
\includegraphics[width=0.3\textwidth,angle =270 ]{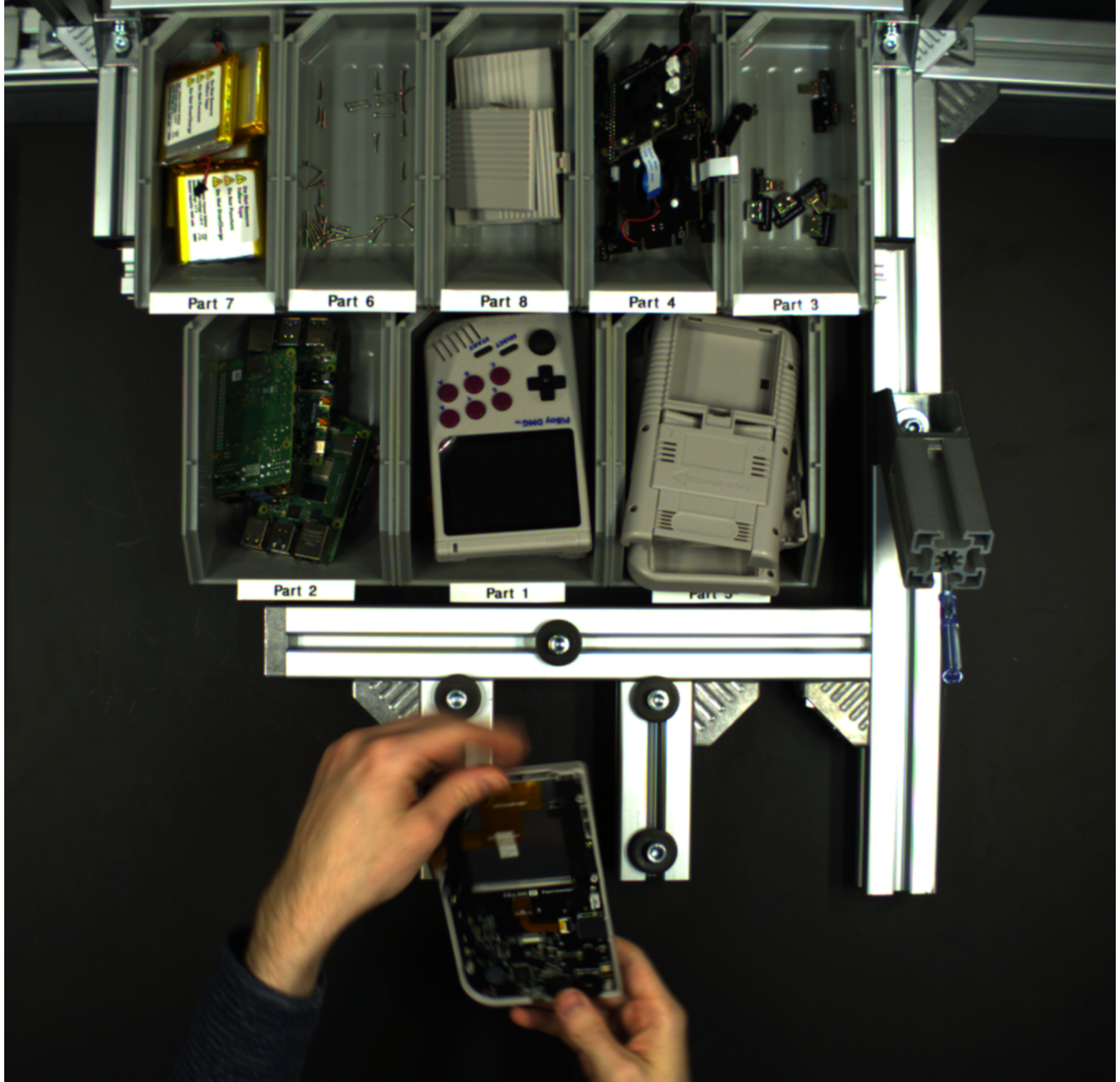}
\caption{Occlusion}
\label{fig:occlusion}
\end{figure}

\begin{figure}
    \centering
    \begin{subfigure}{0.3\textwidth}
    \centering
    \includegraphics[width=\textwidth]{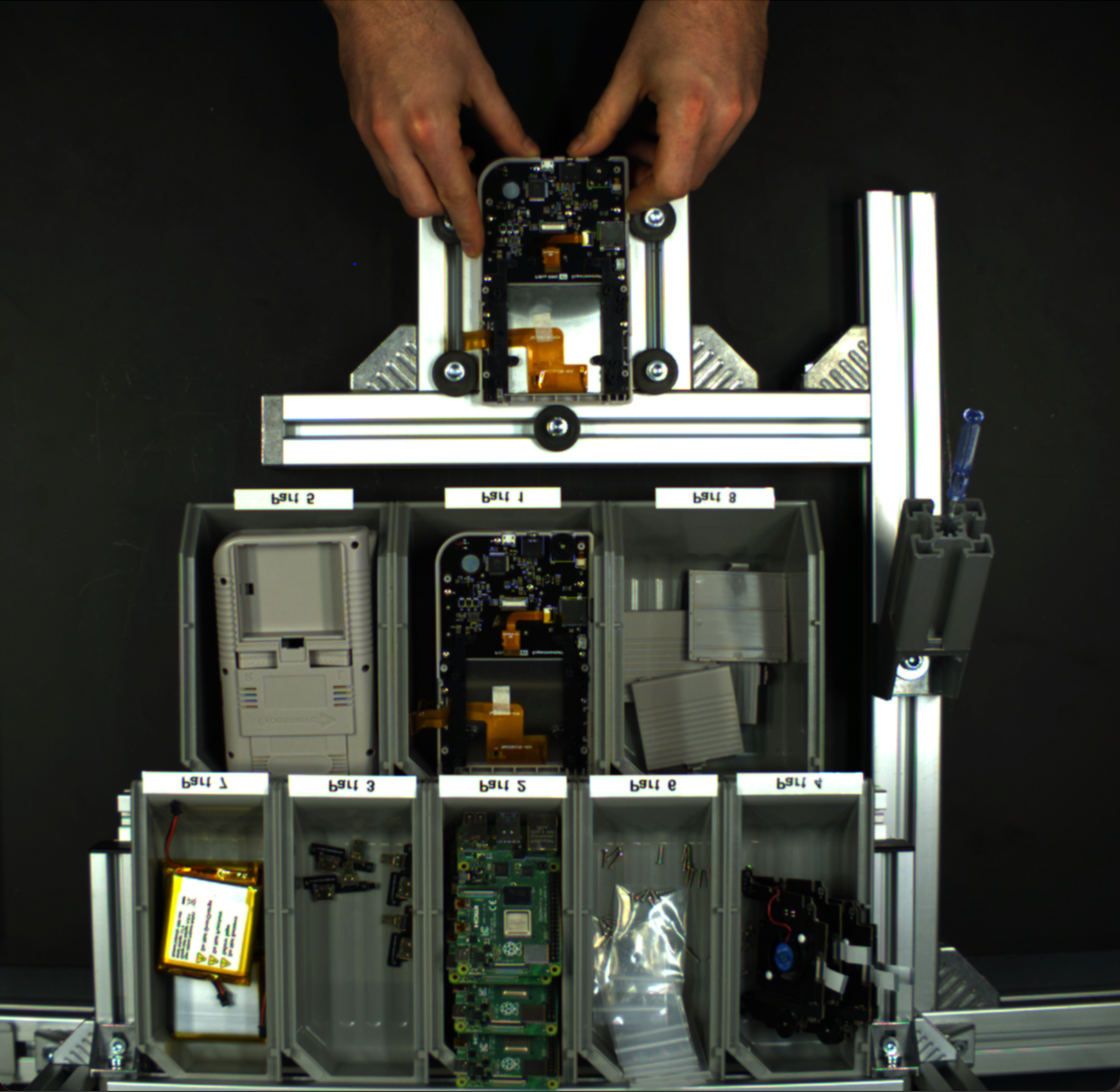}
    \caption{Vertical Flip}
    \label{fig:verticalflip}
    \end{subfigure}
    \hfill
    \begin{subfigure}{0.3\textwidth}
    \centering
    \includegraphics[clip, trim=0cm 0cm 0cm 0cm,width=\textwidth]{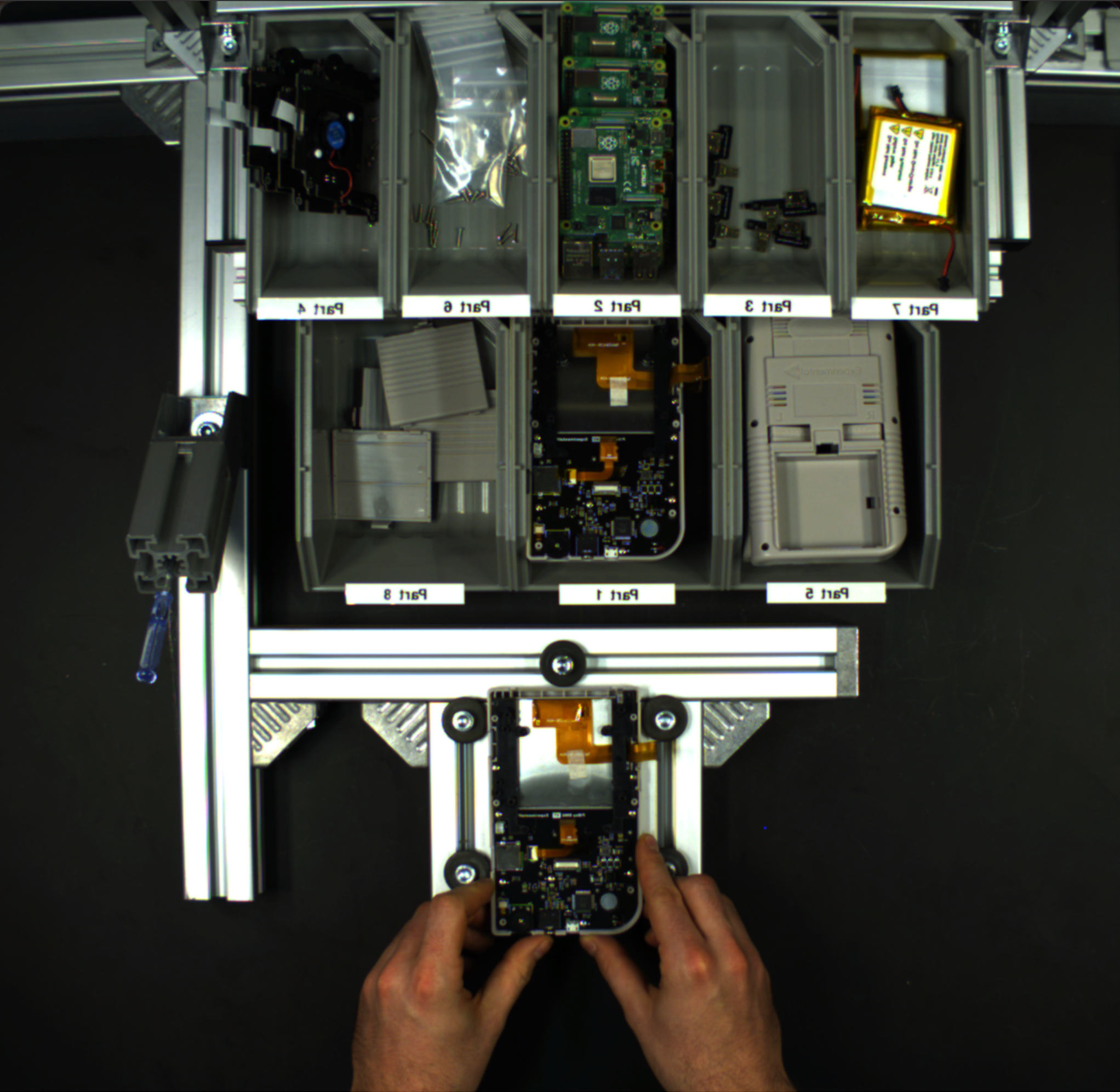}
    \caption{Horizontal Flip}
    \label{fig:horizontalflip}
    \end{subfigure}
    \hfill
    \begin{subfigure}{0.285\textwidth}
    \centering
    \includegraphics[width=\textwidth]{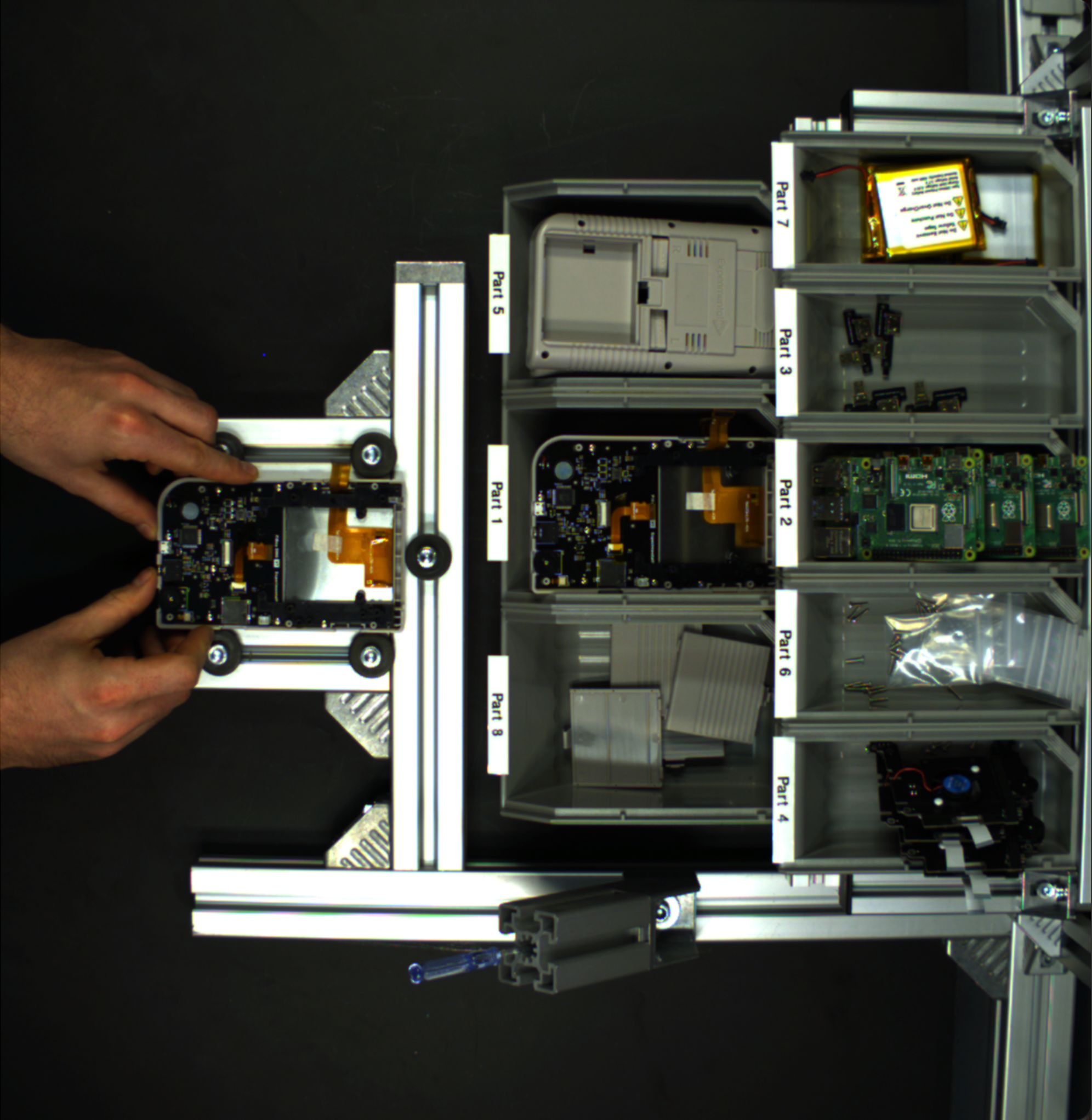}
    \caption{Random Rotation}
    \label{fig:randomrotation}
    \end{subfigure}
\caption{Spatial Augmentations}
\label{fig:SpatialAugmentation}
\end{figure}

\begin{figure}
\centering
%\vspace{0.35\textwidth}
\includegraphics[clip,trim=0cm 0cm 0cm 0cm,width=0.7\textwidth,angle=270]{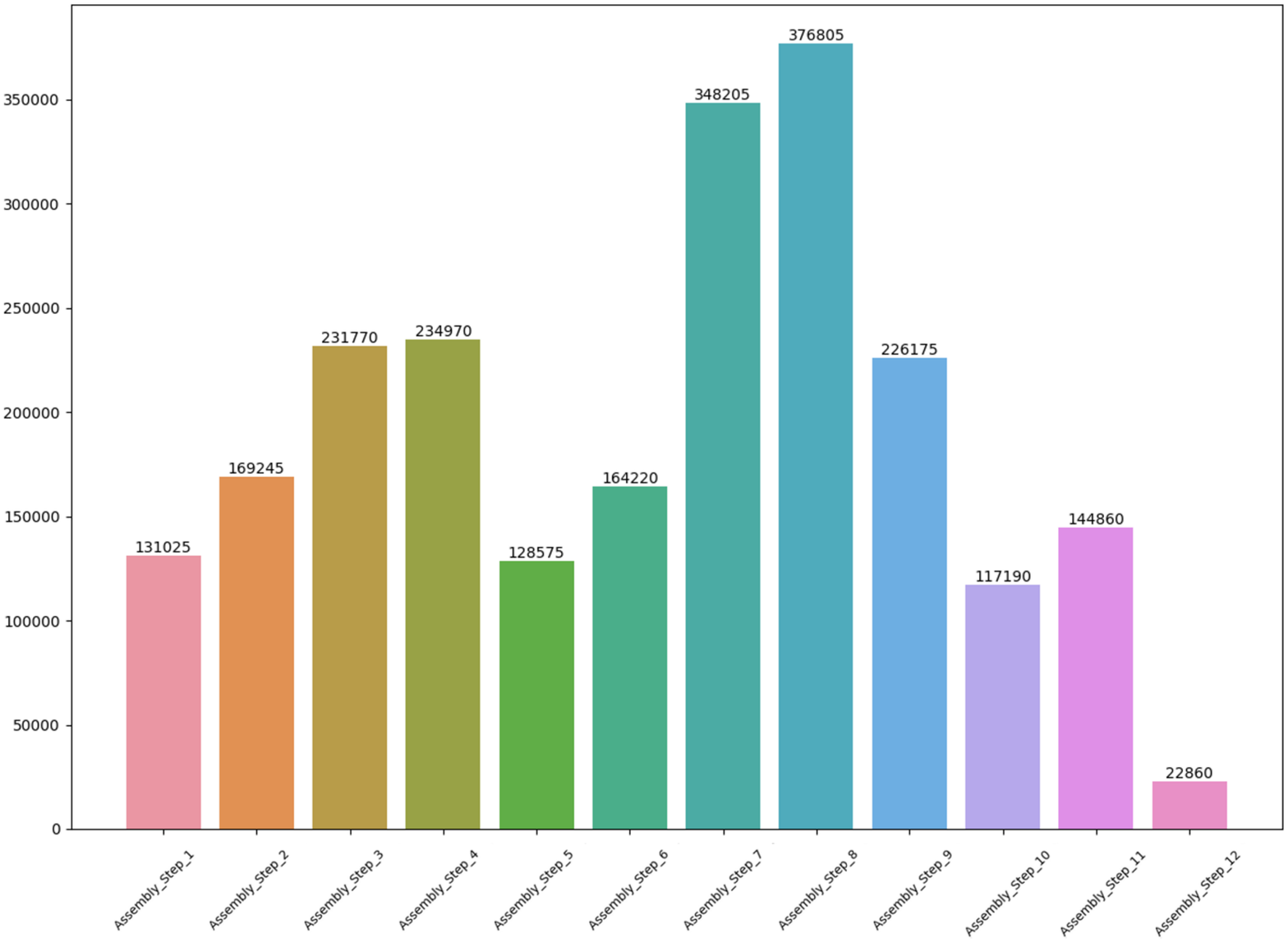}
\caption{Augmented Class Distribution}
\label{fig:augmented_class_distribution}
\end{figure}

This first version of the dataset consists of 12 classes, see Table \ref{table:assemblyclasses}, more precisely actions. 10 actions must be performed by the assembler by hand, and two actions must be performed with the help of a screwdriver which is stored on the right side of the mounting point, see Fig. \ref{testbenchmount}. The tasks were selected according to the frequency as they are also performed in a real world assembly scenario after detailed observation in a real world assembly line. It was found that such real world actions mainly consist of gripping, which needs to be done by nearly all classes to get the respective part, placing (Assembly\_Step1,2,4,5,6,7,8,10,11,12), plugging (Assembly\_Step3,5,9) and screwing (Assembly\_Step7,8) operations.
\begin{table}
  \centering
    \caption{Classes and Corresponding Assembly Actions}
    \label{table:assemblyclasses}
  \resizebox{\textwidth}{!}{\begin{tabular}{ll}
    \toprule
    \cmidrule(r){1-2}
    Classes & Assembly Action\\
    \midrule
    Assembly\_Step1 & Place Housing (Part1) in Mounting Bracket\\
    Assembly\_Step2 & Insert Raspberry Pi (Part2) in Middle of Housing (Position1)\\
    Assembly\_Step3 & Insert HDMI Connector (Part3) in Housing Connectors on the Right (Position2)\\
    Assembly\_Step4 & Place Fan (Part4) on top (Position3) of Raspberry Pi (Part2)\\
    Assembly\_Step5 & Connect Cable with Fan\\
    Assembly\_Step6 & Close Housing (Part5) (Position4)\\
    Assembly\_Step7 & Screw (Part6) in Left Center Housing Screw Hole (Position5)\\
    Assembly\_Step8 & Screw (Part6) in Right Center Housing Screw Hole (Position6)\\
    Assembly\_Step9 & Plug in Battery Cable\\
    Assembly\_Step10 & Place Battery (Part7) into Battery Compartment (Position7)\\
    Assembly\_Step11 & Place Battery Cover (Part8) onto Battery (Position8)\\
    Assembly\_Step12 & Place Finished Product within the Marked Area to the Left\\
    \bottomrule
  \end{tabular}}
\end{table}
A total of 459,180 images were stored for further use. The frames per class are uneven distributed depending on the length and duration, between \raisebox{-0.9ex}{\~{}}5-15 secs, of each task done by each worker on the test bench.
\begin{figure}
\centering
\includegraphics[width=0.73\textwidth,angle=270]{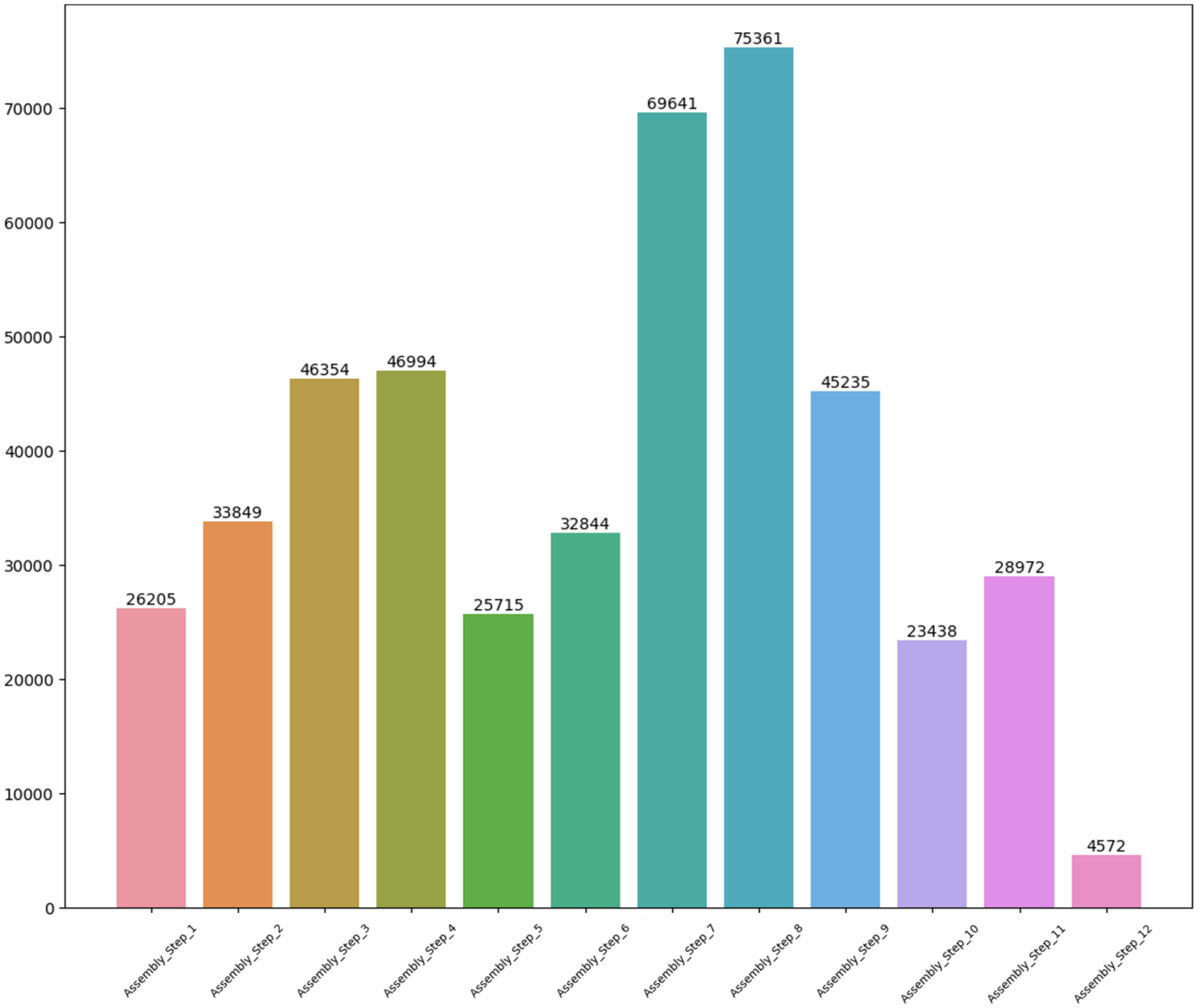}
\caption{Basic Class Distribution}
\label{fig:class_distribution}
\end{figure}
The bins of the necessary components for the demo assembly process are periodically changed in their position between the respective products to increase the variance of the movements. 
As usual in industrial environments, the hands of the worker are also partly occluded by bigger parts, e.g., Part 1, Part 4 and Part 5. An example can be seen in Fig. \ref{fig:occlusion}.
For a better scalability and a higher variance in the dataset, the frames were additionally spatial augmented before the extraction of the keypoints to get more bias into the handedness and the direction of movements. Each sequence was, for the length of duration per task, five times augmented by a random spatial augmentation in vertical flipping which increases the amount of left and right handedness, example can be seen in Fig. \ref{fig:verticalflip}, horizontal flipping in Fig. \ref{fig:horizontalflip} and random rotation by 90 degree in Fig. \ref{fig:randomrotation}, with a probability of 50\% each, to get 2,295,900 frames for the final model training, see in Sect. \ref{Appendix} Fig. \ref{fig:augmented_class_distribution} for the final class distribution. 
If a task in a sequence is shorter than the full duration, zero padding was performed for the pending frames of the sequence for later training on the skeleton information of the hands.
\section{Model Training}
\label{Modeltraining}
In order to confirm that the generated dataset enables the training of a deep learning model, the architecture as well as the training environment are explained in the following sections.
\subsection{Model Training Architecture}
\label{ModelTrainingArchitecture}
\begin{figure}
    \centering
    \begin{subfigure}{.5\textwidth}
    \centering
    \includegraphics[width=.65\linewidth,angle=270]{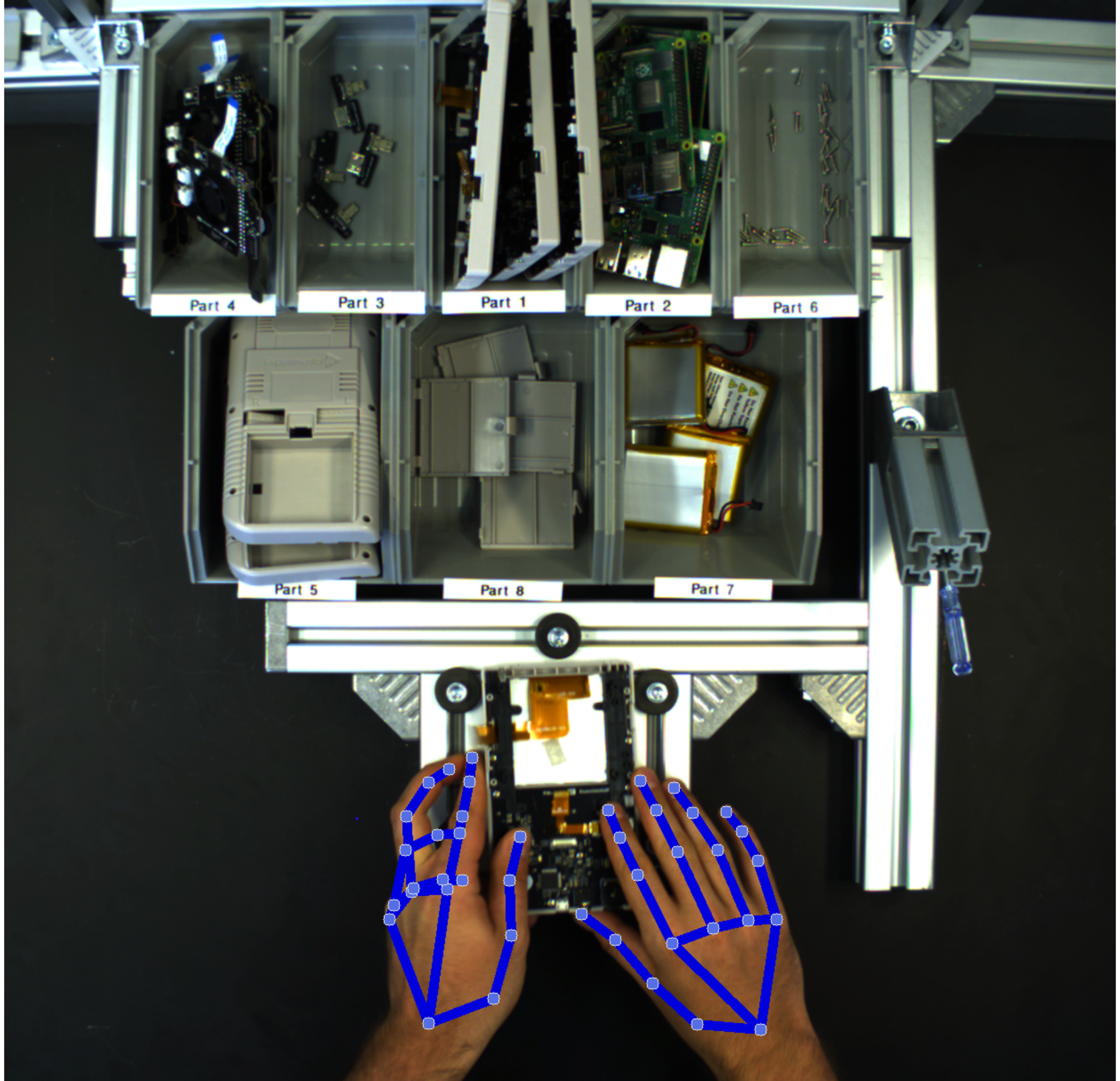}
    \caption{Skeleton Example}
    \label{fig:skeleton}
    \end{subfigure}%
    \hfill
    \begin{subfigure}{.5\textwidth}
    \centering
    \includegraphics[width=.65\linewidth,angle=270]{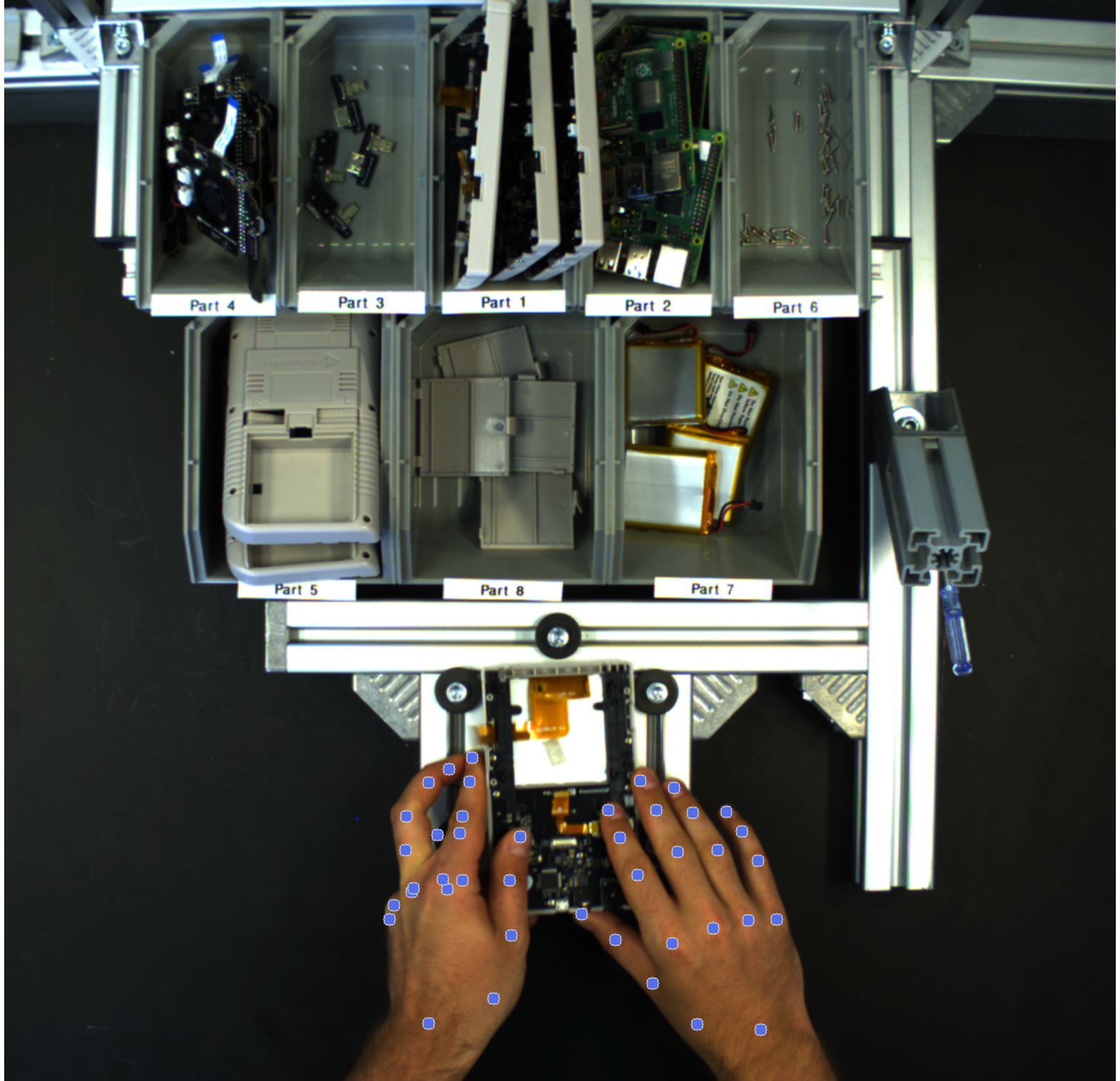}
    \caption{Keypoint Example}
    \label{fig:keypoints}
    \end{subfigure}
\caption{Hand Detector}
\label{fig:handdetector}
\end{figure}

Since the focus of \ac{HAR} relies in this work on especially fine-grained human hand actions, the network architecture is separated into two parts. The first part is the extraction of skeleton features from the hand, see Fig. \ref{fig:skeleton}, which are relevant for human hand actions. More specifically, a keypoint detector to detect human hands and estimate hand posture. In this case, Googles MediaPipe Hands solution is used \cite{Zhang2020}. The output consists of a concatenation per hand, in this case two hands, with 21 keypoints each, see Fig. \ref{fig:keypoints}. Since the keypoint information is provided in 3D world coordinates, x, y and z, an array consisting of 126 data points is created.
This results into the second part of the architecture to classify the sequential correlation of these previous extracted keypoints by an adapted Gated Transformer Network from \cite{Liu2021}. This architecture is based on a two-tower Transformer, where the encoder in each tower capture time-step-wise and channel-wise attention. To merge the encoded feature of the two towers, a learnable weighted concatenation is used as a gate for the final fully connected layers. With this approach, they achieved state-of-the-art results on 13 multivariate time series classification tasks in the domain of Natural Language Processing, but also \ac{HAR} \cite{Liu2021}. Consequently, this work can be classified not only in the area of \ac{HAR}, but also in the area of multivariate time series classification.
\subsection{Model Training Environment}
\label{ModelTrainingEnvironment}
As already mentioned, the Model was created in PyTorch and stacked on top of Googles framework MediaPipe Hands, the training and hyperparameter tuning was done in Microsoft Azure on a STANDARD\_NC6 with 6 vCPUs and 56 GiB Memory. The final model training was done on a GPU which corresponds to half a K80 card with 12 GiB, and a maximum of 24 data disks and 1 NCiS in a duration of 1 hour and 55 minutes.
\newpage
\section{Results \& Examination of the Suitability}
\label{Results}
\begin{figure}
\centering
%\vspace{0.35\textwidth}
\includegraphics[clip,trim=3cm 0cm 3.2cm 0cm,width=\linewidth]{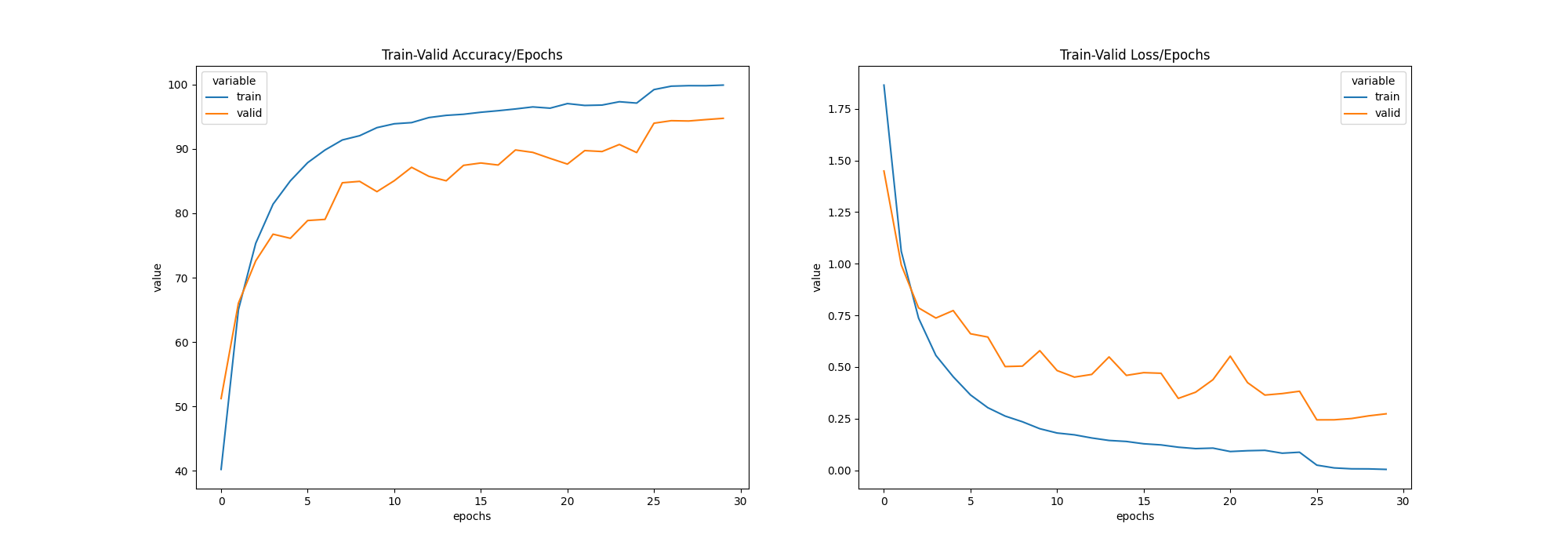}
\caption{Train-Validation Curve}
\label{fig:Trainingcurve}
\end{figure}

\begin{figure}
\centering
\includegraphics[width=\textwidth]{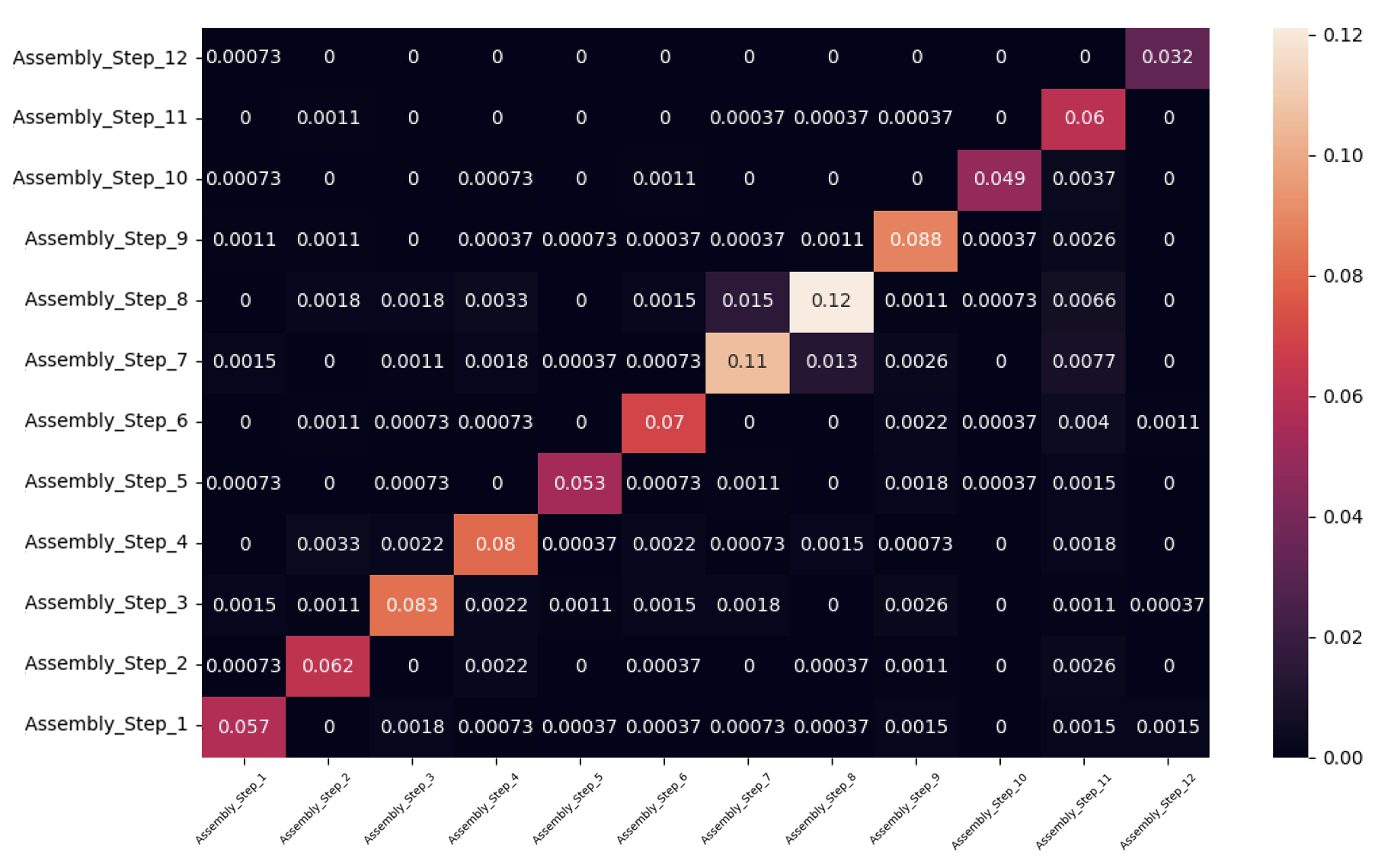}
\caption{Confusion Matrix}
\label{fig:Confusionmatrix}
\end{figure}

\begin{figure}
\centering
\includegraphics[clip,trim=2.5cm 0cm 2cm 0cm,width=0.5\textwidth,angle=270]{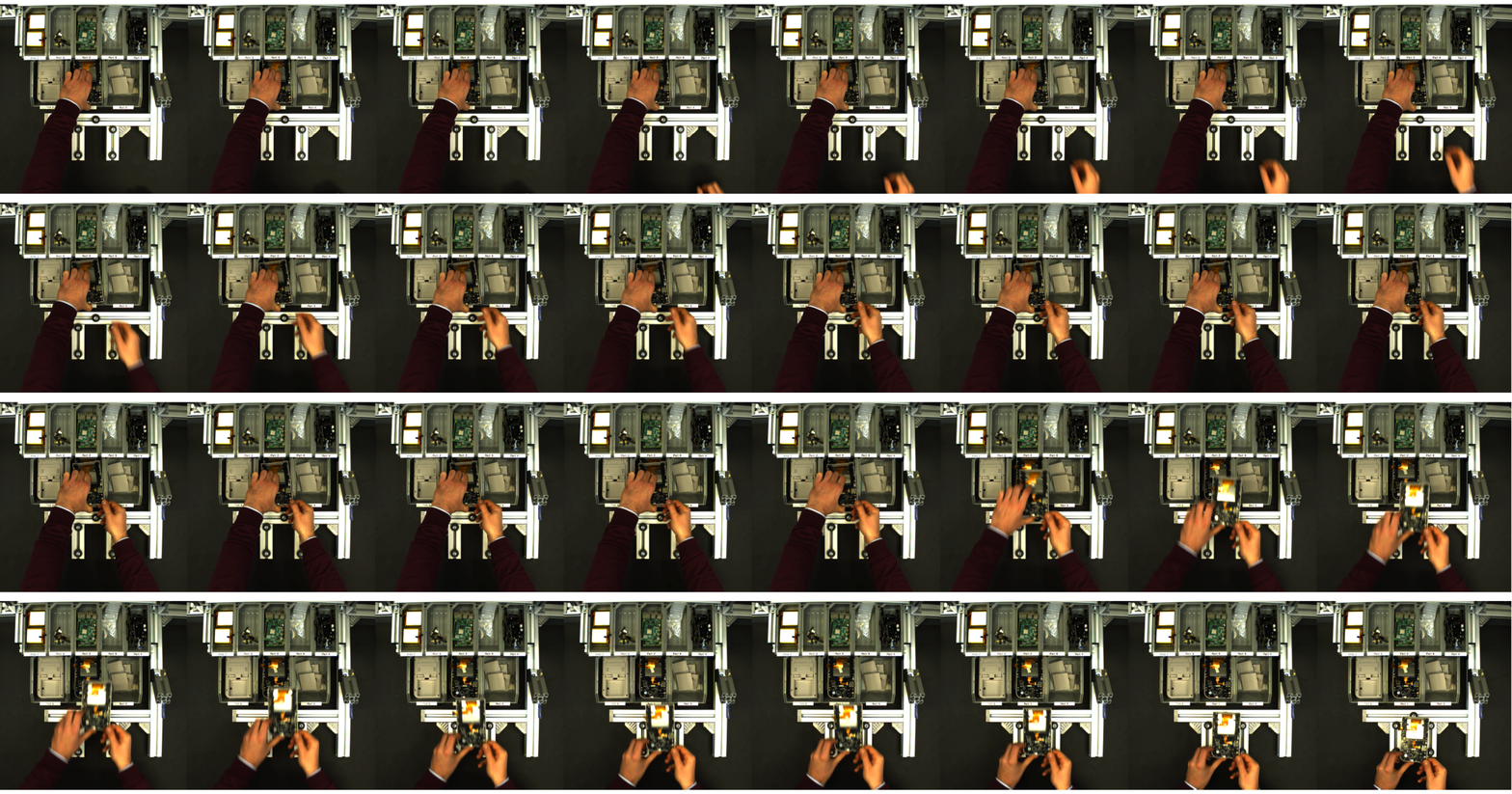}
\caption{Example Assembly Step 1}
\label{fig:Assembly_step_1}
\end{figure}

\begin{table}
\caption{Industrial Hand Action Dataset V1}
\label{table:results}
\centering
\resizebox{\textwidth}{!}{\begin{tabular}{ccccccccc}
\toprule
\cmidrule(r){1-9}
\makecell{Hand-Object\\Interaction} & Classes & Frames & Resolution & FPS & Activity & Environment & Views & \makecell{GDPR\\Constrains}\\
\midrule
Yes & 12 & 2,295,900 & 1936×1216 & 24 & \makecell{Industrial\\Assembly} & Industrial & 1 & No\\
\bottomrule
\end{tabular}}
\end{table}
The workbench created specifically for the dataset, Fig. \ref{testbenchfront}, for simulating industrial assembly tasks is ideally suited for recording the desired work steps and can easily be expanded in the future for further modifications and recording devices. Due to the clear and unambiguous structure of the product which has to be assembled, it is possible to create a suitable ground truth of the respective tasks without being disturbed by the background environment. Although the model trained on this dataset is prone to overfitting due to the size of the network with 18,269,959 trainable parameters correlated to 2,295,900 frames with 126 data points each, which will be investigated in more detail in later experiments, first training results were promising, with a test accuracy of 86.25\% before hyperparameter tuning, and a validation accuracy of 94.73684\% as can be seen in Fig. \ref{fig:Trainingcurve}.
Occlusions in the recorded fine-grained motions, especially during the assembling of bigger parts, are particularly relevant to the recorded dataset and help make the model being trained more robust to unforeseen occurrences. This robustness was further complemented by spatial augmenting. As already mentioned, the distribution of the 12 classes is based on the duration of the respective work steps. It can be seen in Fig. \ref{fig:class_distribution}, that pick and place classes like Assembly\_Step1,2,10,12 were recorded significantly shorter than, e.g., the assembly of the HDMI connector in Assembly\_Step3 or the screwing work in Assembly\_Step7 and Assembly\_Step8. Further, in Fig. \ref{fig:Confusionmatrix} it is visible that most of the tasks are clearly distinguishable from each other. There are overlaps in the respective screw movements, which must be performed on the one hand at the similar position and on the other hand with the same objects.
The results from a final version of a recorded sequence is presented in Fig. \ref{fig:Assembly_step_1}.
In a future version of this dataset, more industrial conditions like gloves and more fine-grained assembly tasks will be included. For a higher scalability, multi camera views are also planned. Further, a brighter diversity of assembly tasks with different hand colors and genders will also be recorded. Additionally, results of the model training will be presented, and the dataset will be published to the scientific community.
\section{Conclusion}
\label{Conclusion}
The availability of a dataset that reflects the real world is an important criterion when investigating the scalability of a deep learning model. The scalability is a fundamental evaluation characteristic, especially in industrial applications. In this work, different image and video datasets similar to industrial assembly applications were investigated and their technical vulnerabilities as well as legal complications, with respect to privacy regulations, are listed. Subsequently, a custom-built dataset corresponding to the presented aspects consisting of 459,180 recorded and after augmentation 2,295,900  frames uneven distributed on 12 typical industrial fine-grained human hand action classes during assembly are presented.
The experiments show that it is possible to train a sequential model with this dataset and achieve promising results. Before the dataset can be published, further experiments need to be performed, e.g., further variance in sequence length and further variation in hyperparameters and model architectures to classify sequences and to avoid overfitting. Since the dataset consists of various fine-grained human actions that occur daily in an industrial environment, this dataset helps test and apply the application of deep learning research methodologies to real world problems, especially in industrial assembly tasks.
\printbibliography
\section{Appendix}
\label{Appendix}

\begin{landscape}
\begin{table}%[htbp]
  \caption{Technical Comparison of Human Action Recognition Datasets}
  \label{table:datasetcomparisontable}
  \centering
  \begin{tabular}{ccccccccccc}
    \toprule
    \cmidrule(r){1-11}
    Dataset & \makecell{Hand-Object\\Interaction} & Modalities & Classes & \makecell{Clips/\\Videos} & Frames & Resolution & FPS & \makecell{Gestures/\\Actions} & \makecell{Hand\\Pairs} & Views\\
    \midrule
    IPN Hand \cite{DBLP:journals/corr/abs-2005-02134} & No & RGB & 14 & 4,000 & 800 & 640×480 & 30 & 13 & 1  & 1\\
    EgoHands+ \cite{EgoHands} & Yes & RGB & 16 & 48 & 130 & 1280×720 & 30 & 4 & 2 & 1\\
    EYTH (EgoYouTubeHands) \cite{aisha} & Yes & RGB & X & X & 1,290 & X & X & X & X & 1\\
    Ikea Assembly Dataset \cite{IKEAASM} & Yes & RGB-D & 4 & 371 & 3,046,977 & X & 24 & 33 & 1 & 3\\
    20bn-something-something-v2 \cite{Goyal2017} & Yes & RGB & 174	& 108,499 & X & X & X & 174 & 1 & 1\\
    Cambridge Hand\\ Gesture Dataset \cite{tensorcanonical} & No & RGB & 9 & 900 & X & X & X & 9 & 0.5 & 1\\
    EGTEA Gaze+ \cite{egtea} & Yes & RGB & 106 & 86 & 14,000 & 1280×960 & 24 & 32 & 1 & 1\\
    MPII Cooking 2 \cite{Rohrbach2016} & Yes & RGB & 59 & 273 & 2,881,616 & 1624×1224 & 29.4 & 59 & 1 & 1\\
    YouCook 2 \cite{Zhou2018} & Yes & RGB & 89 & 2,000 & X & X & X & X & 1 & 1\\
    COIN \cite{Tang2019} & Yes & RGB & 180 & 11,827 & X & X & X & 83 & 1 & 1\\
    Drive\&Act \cite{driveandact} & Yes & \makecell{RGB-D/\\Skeleton} & 83 & X & \raisebox{-0.9ex}{\~{}}9,6Mio. & 1280×1024 & X & X & 1 & 6\\
    WorkingHands \cite{workinghands} & Yes & RGB-D & 13 & X & X & 1920×1080 & 7 & 13 & 1 & 1\\
    ChaLearn Iso/ConGD \cite{chalearn} & No & RGB-D & 249 & 22,535 & X & X & X & 47,833 & 1 & 1\\
    LTTM Senz3D \cite{lttm1}\cite{lttm2} & No & RGB-D & 11 & 1320 & X & X & X & 11 & 0.5 & 1\\
    Multi-Modal Hand\\ Activity Video Dataset \cite{multi-modal_hands} & Yes & \makecell{Thermal/\\RGB-D} & 15	& 790 & 401,765 & X & X & X & X & 1\\
    EgoGesture Dataset \cite{Zhang2018} & No & RGB-D & 83 & 24,000 & 2,953,224 & 320×240 & X & 24,161 & 1 & 1\\
    Gun71 \cite{rogez} & Yes & RGB-D & 71 & X & 12 & X & X & X & X & 5-6\\
    \bottomrule
    X = Information not specified in reference
  \end{tabular}
\end{table}
\end{landscape}
\end{document}